# CENTRAL KURDISH MACHINE TRANSLATION: FIRST LARGE SCALE PARALLEL CORPUS AND EXPERIMENTS

Zhila Amini

Hamedan University of Technology, Hamedan, Iran, zhilaaminirs@gmail.com

Mohammad Mohammadamini

Laboratoire Informatique d'Avignon (LIA), Avignon University, Avignon, France, mohammad.mohammadamini@univ-avignon.fr

Hawre Hosseini*

Laboratory for Systems, Software and Semantics (LS3), Ryerson University, Toronto, Canada, hawre.hosseini@ryerson.ca

Mehran Mansouri

Allameh Tabataba'i University, Tehran, Iran, mehranmansouri9191@gmail.com

Daban Q Jaff

Department of English, Faculty of Education, Koya University, Koya KOY45, Kurdistan Region – F.R. , Iraq, daban.jaff@koyauniversity.org

While the computational processing of Kurdish has experienced a relative increase, the machine translation of this language seems to be lacking a considerable body of scientific work. This is in part due to the lack of resources especially curated for this task. In this paper, we present the first large scale parallel corpus of Central Kurdish-English, Awta, containing 229,222 pairs of manually aligned translations. Our corpus is collected from different text genres and domains in an attempt to build more robust and real-world applications of machine translation. We make a portion of this corpus publicly available in order to foster research in this area. Further, we build several neural machine translation models in order to benchmark the task of Kurdish machine translation. Additionally, we perform extensive experimental analysis of results in order to identify the major challenges that Central Kurdish machine translation faces. These challenges include language-dependent and -independent ones as categorized in this paper, the first group of which are aware of Central Kurdish linguistic properties on different morphological, syntactic and semantic levels. Our best performing systems achieve 22.72 and 16.81 in BLEU score for Ku→EN and En→Ku, respectively.

## 1 INTRODUCTION

Existence of corpora, especially parallel ones, is crucial for the purpose of machine translation. Such resources are useful in the transition from rule-based and dictionary-based machine translation to statistical and neural approaches to this task which are less language-dependent [12][13][11]. The need for such corpora is even more dire in the case of languages which have not benefited from NLP practitioners and researchers developing and formalizing linguistically inspired rules and heuristics. Despite having an abundant number of speakers, Kurdish language still remains a low resource language which has not been subject to extensive computational processing. This is in part due to the variety in its dialects; however, the main reason might be the fact that Kurdish is not the official language of states (except for Iraq, only recently). This has resulted in a lack of funding and/or initiative for NLP research [24]. Recently, there have been attempts by NLP researchers to create resources as well as to perform NLP research on this language [24][23][1][7]. However, to the best of our knowledge, there is no relatively large-scale parallel corpus for any dialect of this language. The only parallel corpus for Central Kurdish-English is that of Ahmadi et al. [2]. In this work, along with two other parallel corpora, a parallel corpus of English-Central Kurdish that only contains 650 translation pairs is presented. This has resulted in a lack of significant research on Kurdish NLP in general and Kurdish machine translation, in particular.

---

* Corresponding author

The objectives of this paper are two-fold: (1) constructing the first relatively large-scale parallel corpus for Central Kurdish machine translation; and, (2) endeavoring to identify the major challenges that Kurdish machine translation faces. Therefore, we construct the Awta[1] corpus containing various genres of texts from different domains. This corpus contains 229,222 sentence pairs and 2.1 M and 2.2 M tokens in the Kurdish and English corpora, respectively. Statistics of the corpus are elaborated on in Section 3. In order to foster research for Central Kurdish machine translation, we publicly share a portion of the Awta corpus containing 100 K translation pairs, randomly chosen while respecting the topic ratio of our corpus. This sample and other resources can be found at the project's web page[2]. In addressing the second objective, we perform experiments in order to benchmark the task of machine translation for Central Kurdish. We build various neural machine translation models with our data and using its different domain-based subset and evaluate the obtained results to extract insights. Our best performing models achieve 22.72 and 16.81 in BLEU score for Ku→En and En→Ku, respectively. Additionally, we design experiments in order to identify the challenges that Central Kurdish machine translation faces as related to its linguistic, inherent properties. Such challenges are observed on morphological, syntactic and semantic levels. In doing so, we perform a comprehensive error analysis of our system's output and provide linguistically inspired details for each. In short, the major challenges that Central Kurdish faces include data sparsity and agglutinative nature of the language. There are other sources of error based on the BLEU score as a metric; such errors do not result in a wrong translation, yet decrease performance statistics. The contributions of this work can be enumerated as follows:

(1) We propose the first relatively large-scale parallel corpus for Central Kurdish-English;
(2) We carefully categorize corpus data into text genres and domains;
(3) Various models are built for Central Kurdish machine translation and detailed statistics are provided; and,
(4) We perform an explanatory error analysis of our system in order to shed light on nuance of Kurdish machine translation.

The rest of this paper is organized as follows. In section 2, we draw upon the relevant body of knowledge to our study. In Section 3, the corpus building process as well as the methodology for building models is elaborated on. This is followed by the evaluation of the systems and analysis of the obtained results in Section 4. In Section 5, we provide an in-depth error analysis of our systems. We conclude our paper by identifying future directions of work.

---

[1] Awta in central Kurdish means 'synonym' or 'equal'
[2] https://github.com/mihemmed/Kurdish-Machine-Translation-Project-KMTP-1

## 2 PRIOR ART

Relevant literature to this work falls into two major categories. The first includes works that attempt to build corpora for machine translation, especially more recent attempts and in particular corpora for low resource languages. These works are particularly important as they contain insights into building proper corpora that enable research as well as real world applications of machine translation. The second category encompasses works that focus on Kurdish MT.

Parallel corpora have been developed for many languages over the past decades. The most recent examples of corpora that are closely related to our work and/or widely used in the literature are Tian et al. [21], Hikaridai et al. [22] and Toshiaki Nakazawa et al. [16]. These parallel corpora contain 15 M, 2 M and 3 M sentence pairs, respectively. The work by Nakazawa et al. [16] describes the first large-size parallel corpus of scientific domain in three languages (Japanese, Chinese and English) including 8 scientific fields. Scientific fields include Medicine, Information, Biology, Environmentology, Chemistry, Materials, Agriculture, and Energy. Another example is the corpus presented by Eisels et al [6] which is available in 6 official UN languages. The data has been collected from the ODS website of the United Nations and it consists of ~81 M sentences in total with the average of 326 M tokens for five out of six official languages. The work by Tian et al. [21] is particularly of interest to us as it uses resources of different genres in order to construct a balanced Chinese-English parallel corpus. Genres include News, Spoken, Laws, Thesis, Education, Science, Subtitle, and Microblog. This corpus, denoted UM-Corpus, contains more than 15 M parallel sentences, a part of which including 2 M pairs has been made publicly available. More importantly, for low resource languages, there have been attempts to build parallel corpora and machine translation systems in recent years [2][8]. The work by Choudhary et al. [4] builds several parallel corpora containing sentence pairs between English and such other low resource Indian languages as Tamil, Malayalam, Telugu, Bengali and Urdu. Their corpus contains the following number of sentence pairs for each of those languages, respectively: ~187 K, ~554 K, ~83 K, ~665 K and ~41 K. Authors in this work introduce the agglutinative nature of those languages and OOV tokens as major challenges of their neural machine translation system. Their system manages to outperform Google translator significantly for Tamil and Malayalam. In the work by Rayson et al. [8], a benchmark evaluation dataset that contains about 10 K Igbo-English sentence pairs is created. Igbo is one of the largest languages of West Africa, spoken by over 50 million people globally. The corpus is mostly extracted from news Domain. In this work no machine translation model has been built. In the work by Duygu Ataman [3], a parallel corpus of Northern Kurdish, Turkish and English containing content from news domain is presented. Number of sentences in Turkish-English, English-Kurdish and Turkish-Kurdish corpora are nearly 35 K, 6.5 K and 7.4 K, respectively. To evaluate the quality of the collected corpora, bilingual and multilingual NMT systems are trained. The BLEU scores of bilingual English-Kurdish and English-Turkish models are 5.41 and 11.5, respectively.

Table 1. Token number of raw data sources categorized by source type and their topics

|  | TED | Politics | Literature | Biography | Philosophy | Scientific | Mixed | **All** |
|---|---|---|---|---|---|---|---|---|
| Number of Pairs | 141,024 | 6,057 | 1,570 | 7,082 | 311 | 6,510 | 66,668 | **229,222** |

Most related to our work are works on Central Kurdish machine translation. There have been very few studies in the field of Kurdish MT due to lack of parallel corpora. The only published or reviewed attempts are Kaka-Khan [10] and inKurdish[3], and most recently Ahamdi and Masoud [1] as well as Ahmadi et al. [2]. None of these attempts prepare parallel corpora of significant size curated especially for the task of machine translation. The only work on building Kurdish parallel corpora is that of Ahmadi et al. [2]. Authors build a corpus of 12,372 pairs between two Kurdish dialects of Central Kurdish (Sorani) and Northern Kurdish (Kurmanji) alongside a small corpus of 1,797 and 650 translation pairs of English-Northern Kurdish and English-Central Kurdish, respectively. As this is the first attempt to build an English-Central Kurdish corpus, we review this work in more depth. The average number of tokens in each sentence is 17 to 20. A statistical MT model has been developed in order to test and evaluate the corpus, where 10% of the corpus has been chosen randomly in order to be used as a test set. The BLUE score of the statistical machine translation for Central Kurdish-English, Central Kurdish-Northern Kurdish, and Northern Kurdish-English corpus are 17.74, 17.08, and 11.06, respectively. In [8], Kaka-Khan provides such tools and resources as morphological dictionary and bilingual dictionaries to develop a rule-based MT Kurdish-English MT system. This system is developed based on the open-source MT framework, Apertium, developed by Forcada et al. [7]. The inKurdish MT system, as reviewed by Kaka-Khan and Taher [19], is a dictionary-based MT system, reportedly the first to offer online translation services. However, this system performs very poorly in dealing with such linguistically complex forms as long sentences, proverbs and idioms. Lastly, Ahmadi and Masoud [1] describe three available Central Kurdish-English domain-specific parallel corpora that are available online though not especially curated for the task of machine translation. They develop two neural machine translation models with different hyper parameters to report the challenges of MT for Central Kurdish.

The available and used parallel corpora are Tanzil Corpus with 92,354 sentence pairs, TED Talks with 2,358 parallel sentences, and Kurd-Net, a lexical-semantic resource with 4,663 definitions. This work uses in-domain data for testing purposes. Additionally, the relatively limited number of translation pairs can result in less generalizability.

Our work extends the current literature in several ways. First and foremost, we provide the first relatively largescale Central Kurdish-English parallel corpus, the Awta corpus, with 229,222 pairs. This corpus is nearly 353 times as large as the existing corpus [2]. Secondly, we build various models and evaluate the model performances extensively. In doing so, we provide qualitative analysis of our NMT systems drawing insights with respect to language-specific properties of Central Kurdish which can be considered for building more robust MT systems.

---

[3] https://www.inkurdish.com/

## 3 PARALLEL CORPUS

In this section, we elaborate on the collection and processing of the Awta parallel corpus. We endeavor to relieve some of the known problems of Central Kurdish digital content, as elaborated on in the following. Additionally, the corpus statistics is presented in detail.

### 3.1 Data Source Gathering

The Central Kurdish-English parallel corpus was constructed from various resources covering the following domains: ~141 K sentence pairs of TED Talks subtitles collected from the web[4], ~15 K sentence pairs extracted from digital books with four different genres after getting permission from authors, and ~73 K sentence pairs gathered from mixed sources, with domains including academic content (Table 1). Our book data includes Biography, Politics, Literature and Philosophy with ~7 K, ~6 K, ~1.5 K and ~0.3 K pairs, respectively. TED Talks have the minimum and books have the maximum average token per line in both Kurdish and English corpora, with 7.24 and 15.53 in the Kurdish corpus and 7.89 and 16.28 in the English corpus, respectively. The final corpus has the average of 12.11 tokens per line in the Kurdish corpus, and 12.55 tokens in the English corpus. The vocabulary size in the Kurdish corpus is about 200 K words which is almost 130 K words more than the vocabulary size in the English. Table 2 shows the general specifications of the parallel corpus. The finalized Kurdish Parallel Corpus consists of about 230 K sentence pairs and over 2 M tokens.

Table 2. General specifications of the Awta parallel corpus

|  | Kurdish | | | | English | | | |
|---|---|---|---|---|---|---|---|---|
|  | **TED** | **Books** | **Mixed** | **All** | **TED** | **Books** | **Mixed** | **All** |
| **Number Of pairs** | 141,024 | 15,020 | 73,178 | 229,222 | 141,024 | 15,020 | 73,178 | 229,222 |
| **Number of Tokens** | 1,021,036 | 241,682 | 866,590 | 2,129,308 | 1,113,449 | 253,714 | 846,978 | 2,214,141 |
| **Vocabulary size** | 100,816 | 46,062 | 71,839 | 218,717 | 32,475 | 24,004 | 34,347 | 90,826 |
| **Average token per pair** | 7.27 | 15.53 | 9.06 | 12.11 | 7.89 | 16.28 | 9.04 | 12.55 |

### 3.2 Content Alignment

In this section, we discuss the alignment process of the Awta corpus and the major challenges we have faced in doing so, including the various solutions and steps we acquired to align files with different formats. Since the corpus raw data has been collected from diverse sources with different formats, the alignment process of each domain of resources was different. The alignment process has been a combination of automatic alignment at paragraph and sentence level where possible followed by a manual review and correction by human annotators. We exploited InterText [25] for the manual alignment. Intertext is a parallel text alignment editor for building bilingual sentence pairs, providing various editing tools and functions such as splitting and merging aligned elements and editing and re-positioning elements. Here, we present the alignment process in more detail.

The alignment process of books involved three main steps. First, paragraphs were aligned one-to-one manually. Second, automatic sentence alignment was applied on each paragraph. Lastly, we checked on the aligned corpus in the InterText one more time to make sure it is all aligned in a one-to-one level. The TED Talks were mostly aligned automatically followed by a review and manual correction of errors by human annotators. There have been many errors arising from non-matching time information. Human annotators have rectified all such instances manually. With respect to the mixed data sources, we first removed all the special tags and also empty lines or lines with no translation. And then, we performed automatic sentence alignment of the content. Finally, we converted all the aligned datasets into plain text and combined them.

---

[4] https://amara.org/en/

### 3.3 Text Normalization and Preprocessing

Normalization is an essential step in order to prepare our corpus for later preprocessing. This is different from model specific preprocessing of textual content that is purposefully done in order to achieve better results. This normalization is to relieve the challenges that Central Kurdish textual (particularly digital) content faces generally. Such challenges are mainly due to the non-standardized typography. We resolved several issues during the normalization procedure of our Kurdish corpus. Firstly, we standardized and unified the script based on widely accepted Unicode codes for Central Kurdish Arabic-based script [3][24]. This step is important as a great portion of the Kurdish corpus contents were written using non-Unicode Kurdish keyboards or Unicode keyboards that use non-Kurdish Unicode characters. Secondly, due to the Central Kurdish typography similarities to Persian and Arabic, non-Kurdish forms of characters happen in the text. As an instance, the character ك and ک both have the same sound which is /k/, whilst the character ك is Arabic and ک is the correct form of the Kurdish character. The Arabic ي which resembles the Kurdish ی is another instance. We unified all such instances. On the phoneme level, the phonemes /r/ and /rr/ as well as /l/ and /ll/ look very similar, distinguished only by a special accent under and above such characters, respectively. The writing forms are as follows: /r/ 'ر', /rr/ 'ڕ', /l/ 'ل', and /ll/ 'ڵ'. These phonemes are incorrectly interchangeably used in Kurdish writing. In order to deal with the aforementioned challenges, we use AsoSoft normalizer [14] which is a very useful tool for normalizing Central Kurdish text.

In what follows, we review the preprocessing steps we have performed before building models. Diversity of writing forms in Central Kurdish is a major challenge that we have to deal with in the preprocessing step. The Central Kurdish language has a non-standardized writing form causing lexical variations based on scripting guidelines as well as subdialect-based variations. The first category of variations happen since there is not a unified set of rules as to how words are written in their different morphological forms. For instance, the verb دامانگرتن 'we downloaded them' can be written as دامان گرتن or دامانگرتن. Additionally, another set of lexical variations happen due to subdialect-based differences in pronunciation as well as formal and informal writing forms. For example, words ناشکێت and ناشکی both mean 'it won't break'; however, the first form has a ت attached to the end of it and usually happens in formal text. Another example is with regards to the imperfective aspect in Central Kurdish, whose formal and informal forms are different. For instance, the word 'should' can be written as دەبێت and ئەبێ, the latter of which is the formal form. This difference can occur in different sub-dialects of Central Kurdish. These words can be replaced by one unique form of the word. Finally, named entities can be written in many different forms since the Central Kurdish is a phonemic language and writing is correspondent to how words are pronounced. We unify instances of such named entities. Further, we resolve the issue of conjunctive و in Central Kurdish [24]. The conjunctive و is an independent Kurdish word which means 'and' in English. It should be written separately, however sometimes the writers join it to the previous word without any white space in between. That is against Kurdish typography rules. For instance, in ژنانو پیاوان 'men and women', the conjunctive و follows the previous word ژنان, while it should be written as ژنان و پیاوان. We used AsoSoft Correction Table [23] with the size of nearly 20,000 pairs of the corrected forms of common errors, extracted from a large text corpus. This has allowed for replacement with the correct and/or a unique spelling of the Kurdish words. Overall, about 92 K words were replaced by their correct font in the Correction Table.

Table 3. Model performance

|  | Joint Training | GLOVE-100 | GLOVE-200 |
|---|---|---|---|
| **En→Ku** | 16.16 | 16.3 | 16.81 |
| **En→Ku** | 21.6 | 21.75 | 22.72 |

Table 4. The size of Kurdish lexicon

| Size | OOV | BLUE |
|---|---|---|
| **50k** | 6% | 16.16 |
| **100k** | 3% | 15.83 |

Table 5. Comparison to Microsoft Bing translator

|  | Our best model | Microsoft Bing |
|---|---|---|
| **En→Ku** | 4.79 | 7.4 |
| **En→Ku** | 6.82 | 13.59 |

Punctuation normalization is the next step. We put spaces before each punctuation except for the apostrophes which are not supposed to be disjoint from the words. We also put white spaces between numbers and words if there was not any, to make sure numbers would be detected automatically. Then, we converted them to words. We also removed all the extra white spaces and the empty lines. Finally, the pairs with more than 50 tokens as well as pairs that are exactly identical on both sides were removed.

## 4 EXPERIMENTAL AETUP AND EVALUATION

### 4.1 Models and Metrics

Here, for the sake of reproducibility, we describe the specifications of our train and test sets, model hyperparameters as well as metrics used for evaluation purposes. The final version of the Awta text corpus used for experiments includes 229,222 sentence pairs. We separated 5000 samples for testing and 5000 samples were used for validation. The remaining part is used to train the NMT system. We prepared a second dataset to evaluate our system on out-of-domain data. We collected sentence pairs posted on a social platform consisting of mainly brief explanations of English-Kurdish news articles. The final version of the aforementioned collected test set contains 1000 sentence pairs.

For all our models, we employ the RNN Encoder-Decoder architecture, implemented in OpenNMT machine translation library [9]. The architecture used in our experiments has two hidden LSTM layers with 512 neurons in the encoder and the same architecture is used in the decoder. The system was trained with the Adagrad optimization algorithm.

In the case of not sufficient data available for training, using pre-trained embeddings with monolingual text corpus can help the training of the DNN based NMT systems [18]. In our experiments we tested both cases: joint training of word embeddings and Encoder Decoders, using pre-trained word embeddings for both source and target languages. On both sides, we used the GloVe word embeddings [18]. For English, the vectors were acquired from the web[5]. We trained the Glove embeddings leveraging the AsoSoft text corpus [23]. The AsoSoft text corpus contains 183 M tokens. In the trained embeddings, all unique tokens are included to cover the maximum number of tokens in the machine translation corpus.

In our experiments we use the BLEU (BiLingual Evaluation Understudy) score which is proposed in [17]. The BLEU metric ranks machine translation sentences by function of same n-grams between the MT translation and the human translation. Because the BLEU score needs the exact matching of words between human and machine translation, for an agglutinative language such as central Kurdish the values of BLEU is lower in comparison to Ku→EN translations. This is discussed in more detail in the qualitative analysis section.

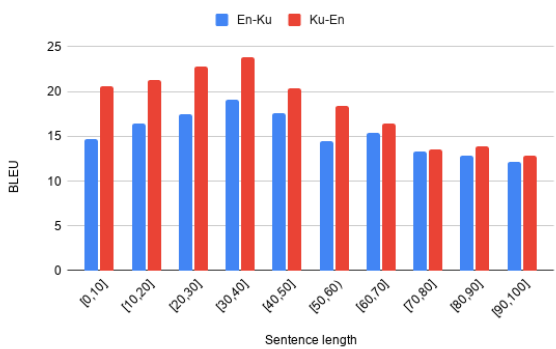

Figure 1. BLEU results based on sentence length

**4.2 Evaluation**

In this section, we present the results obtained by the NMT systems we have built. We build our systems with Kurdish both as target and source language. For such models, we obtain results while varying sentence length, lexicon sizes, embedding dimension sizes, and so forth, in order to draw insights for Kurdish machine translation and help identify error types and reasons, as elaborated on in the next section.

As seen in Table 3, our best models achieve a BLEU of 16.81 for En→Ku translation and 22.72 for Ku→En. We present results obtained with systems where word embeddings are jointly learned with the system as well as with exploitation of pre-trained word embeddings. Pre-trained word embeddings are a common way to improve the performance of NMT systems as they reduce the number of trainable parameters that helps to train the network with a smaller dataset. In our experiments we used Glove word embeddings for both sides (Ku and En). The results show that we have an improvement in terms of BLEU, improving En→Ku performance from 16.16 to 16.81 and Ku→En from 21.6 to 22.72.

---

[5] https://nlp.stanford.edu/projects/glove/

As indicated in Table 3, the performance of En→Ku translation is worse than Ku→En in all of the models. One possible reason for this could be the fact that when Kurdish is the target language, the number of OOV is higher in comparison to Ku→En translation. We increased the size of the lexicon from 50 K to 100 K for Kurdish language. While the ratio of OOV decreases, the overall performance of our system does not improve. The results for this experiment are presented in Table 4. We believe that this lack of improvement arises from two factors. Firstly, lexicon size increase means a huge increase in trainable parameters. Secondly, for the added lexicon, there are not enough training instances for the system to learn the context. This can be viewed as a trade-off between lexicon size and data-sparsity which can be resolved employing larger parallel corpora.

Further, we investigate our best model's performance on test sets built with varying sentence lengths. We present results of this experiment in Figure 1. As can be seen in the figure, there is a visible pattern where the model's performance increases with increasing sentence length up to a certain value; after reaching the peak performance at the (30, 40] bin with a BLEU of 23.8, the model's performance experiences a downfall almost everywhere. The model performs worst on the test set containing the longest sentences. This is in line with literature indicating that the performance of NMT systems for long sentences decreases [17]. An interesting observation is that the performance of the models for both En→Ku and Ku→En converges with increasing sentence lengths.

The performance of the MT translation system is closely related to the domain of training data [11]. In another experiment, we tried to evaluate our best model on an out-of-domain test. We collected 1000 En→Ku sentences that came from new stories. The results obtained by our system reduce dramatically in comparison to the in-domain test set. We did the same experiment with Microsoft Bing Translator, the only available commercial MT system for the Kurdish language. However, the data used to train Microsoft models is not documented and we don't know if the news data is used in their training data. The results for this experiment are presented in Table 5.

## 5 QUALITATIVE ERROR ANALYSIS

Apart from the methodological approaches to machine translation and the challenges due to data sparsity for training such systems, translating between a language pair involves difficulties that have to do with linguistic complexities. The challenges that machine translation systems face can be regarded as language-dependent and -independent ones. Language independent challenges usually involve universal linguistic properties, occurring in almost any language. For instance, which synonym should the system choose for translating a certain token based on the context, if the synonyms have different connotations? On the other hand, translating between languages that have different sentence structures or morphological systems can be challenging, where MT systems can find it difficult to capture these intricacies. In this section, we perform a detailed error analysis of our MT system endeavoring to achieve a better understanding of our system's performance. Also, we aim at recognizing major language-dependent challenges Central Kurdish MT systems face. A summary of such error types is presented in Table 6. As can be seen, we categorize our system's major sources of error into lexical, semantic and syntactic types. Each of these types have some subtypes. While these challenges affect system's performance based on the evaluation metric (here BLEU), sometimes the translated sentence conveys the source sentence's meaning perfectly. We identify such cases where the difference is due to the Kurdish language properties.

Lexical errors consist of those that arise from the Kurdish writing forms practiced by authors. As mentioned earlier, Although Kurdish language academy, writers and publishers have made several attempts to standardize Kurdish language writing, there is yet to be a widely accepted set of regulations. In [23][14], several aspects of Kurdish language are discussed that cause lexical variations. Dialect based variations, loan-words and foreign names are among the common sources of lexical variations that caused errors in the current study. We categorize such errors as follows:

- *Writing Regulations*: Central Kurdish is an agglutinative language which results in numerous morphological forms. For such a language and considering its Arabic-based script, each morphological form can be written is various ways with regards to joining components of the morphological components. For instance, the word دایانگرت (they downloaded it) can be written as دایان گرت and دایانگرت. While there is not a widely accepted set of regulations, in many cases there are conventional procedures. However, a great body of writers of Central Kurdish do not follow such conventionally accepted set of regulations. Such variations do not harm the meaning, while they result in a lower BLEU score. Post-processing can be done in the pipeline of Central Kurdish MT systems in order to uniform such cases. However, such components require NLP tools which are yet to be developed and widely accessible.

- *Subdialect-based Variation*: Due to lack of standardization in writing, words can be pronounced and written differently by speakers of different dialects [15]. This happens more often in the case of named entities as well as loan-words. For example, foreign named entities are pronounced and written differently by speakers of different dialects. Take 'Switzerland' for an example, which has at least two written forms of سویس and سویسرا in Central Kurdish. It is probable that speakers who write Switherland in those forms have borrowed it from Arabic and Persian, respectively. Another example is the long /u/ versus short /u/; speakers of different regions pronounce words containing such a letter differently. The long /uu/ is written as وو in Kurdish, while the short /u/ should be written as ۇ; however, majority of Kurdish writers are not aware of this difference. Therefore, errors like the following are abundant: سودان /sudan/ is translated into سوودان /suudan/ by the machine, and while the reader understands it, this affects the performance negatively.

Semantic category of error types encompasses those errors that arise from morpho-semantic properties of language and, sometimes, of Kurdish language, specifically.

- *Synonymy*: This type of error refers to those where the system outputs a correct synonym of a token instead of our gold standard human translation. Such errors can happen in any language translation system with any given pair of languages. As an example, the word پیاو in the sentence ئەو پیاوەی جانتاکەی پێ بوو can be translated as 'man' and also 'guy'. In our system, a considerable number of errors can be attributed to this challenge.

- *Named Entities*: Named entities are among common out of vocabulary (OOV) errors in MT. Automatic transliteration as a post editing technique, can improve the quality of MT systems. The automatic detection of named entities in Central Kurdish is a serious challenge as they cannot be signified by uppercase letters similar to English language. Therefore, the integration of NER systems with NMTs can help to improve the quality of En→Ku machine MT systems.

Lastly, the syntactic type of errors includes those which arise from Kurdish language syntactic properties such as sentence structure, inflections, clitics and so forth. Such syntactic properties which can affect the MT system's performance drastically are described in the following.

Table 6. Summary of most common error types

|  | Number | Type | Instance | | |
|---|---|---|---|---|---|
|  |  |  | **Source** | **Human** | **Machine** |
| **Lexical** | 1 | Writing Regulations | They downloaded it | دایان گرت | دایانگرت |
|  | 2 | Subdialect-based Variation | Switzerland | سویس | سویسرا |
| **Semantic** | 3 | Synonymy | ئەو پیاوەی جانتاکەی پێ بوو | The guy with the suitcase | The man with the suitcase |
|  | 4 | Named Entities | When Hamid arrived in Lausanne, he went to meet Vaziri at his home | کاتیک حامید گەیشتە لۆزان چوو تا وەزیری لە ماڵەکەی خۆی ببینێت | کاتیک حامید گەیشتە لۆزان چوو بۆ دیتنی وەزیری |
| **syntactic** | 5 | Prodropness | We need to compensate for the fluid loss | ئێمە دەبێت قەرەبووی لە دەست دانی ئەو شلەمەنییە بکەینەوە | دەبێت قەرەبووی لە دەست دانی ئەو شلەمەنییە بکەینەوە |
|  | 6 | Agreement | Both Alan and I have been very interested in trying to relieve human suffering | من و ئالان هەر دووکمان خەمی ئەوەمان بوو کە ژانەکانی مرۆڤایەتی کەم بکەینەوە | ئالان و من زۆر حەزم لە هەوڵدان بوو بۆ ئەوەی و ئازاری مرۆڤەکان کەم بکەمەوە. |
|  | 7 | Word Order | healthy urban planning | پلاندانانی تەندروستی شاری | پلاندانانی شاری تەندروست |
|  | 8 | Izafe | Sleep can often be impacted by past trauma | زۆربەی کاتەکان خەو دەتوانێت بچێتە ژێر کاریگەری ترۆومای رابردوو | زۆربەی کات خەو دەتوانێت لە ژێر کاریگەری ترۆومای رابردوو دا بێت |

- *Prodropness*: Kurdish is one of the prodrop languages in which subjects can be eliminated in independent and subordinate clauses, and the information about person and number of this subject pronoun, can be represented through agreement marker in verbs. For instance, 'we need to' in the sentence 'we need to compensate for the fluid loss' can be translated as ئێمە دەبێت. However, it also can be translated as دەبێت in which the subject ئێمە is dropped. The person information is added as an inflectional affix to the verb.
- *Agreement*: Subject Agreement in transitive verbs which are made from past stem is fulfilled through enclitic. Such an agreement is fulfilled via a clitic, representing the subject, added to the object. Additionally, the verb represents the person and number in Kurdish. Sometimes the system cannot capture these agreement structures. As an instance, the system cannot identify the number of subjects

خەمی ئەوەمان بووە کە ژانەکانی مرۆڤایەتی کەم and instead of recognizing the verb in the first person plural بکەینەوە, the system recognizes it as first person singular بۆ ئەوەی ئازاری مرۆڤەکان کەم بکەمەوە.

- *Word order*: There are many linguistic components whose order of appearance are different than English. Such components include preposition, circumfix, noun head, relative clause, sequence of adjectives, sequence of adverbs and adjectives of degree. The word order flexibility in Kurdish is very high. Besides, in terms of case inflection and verbal inflection, Kurdish has a lot of equipment. Sometimes, the translation system cannot identify correctly the sequence of nouns and adjectives which come together and identifies them conversely and incorrectly. For example, instead of پلاندانی شاری تەندروست as the translation of 'healthy urban planning', system outputs پلاندانی تەندروستی شاری.
- *Izafe*: Izafe Is a grammatical particle which connects words or phrases together. In Kurdish, izafe particles occur between the head and its dependent, which can have different categories and functions. For example, head can be nominal, adjectival or adpositinal. Therefore, izafe in kurdish has three types of allomorphs one of which appears based on definiteness and phonology conditions. Our system sometimes cannot identify the izafe marker. For instance, کاریگەری ترۆوماى رابردوو should include the izafe marker ی, resulting in the correct form کاریگەریی ترۆوماى رابردوو.

## 6 CONCLUDING REMARKS AND FUTURE WORK

In this work, we presented the Awta Central Kurdish-English parallel corpus containing 229,222 translation pairs. The Awta corpus contains content from several various genres. We publicly share 100 K translation pairs in an attempt to foster research in this field. We also designed and performed experiments in order to understand the intricacies of Central Kurdish machine translation. In doing so, we performed extensive exploratory analysis of errors made by our system. Future work can encompass several lines of work including the following. Firstly, a larger corpus including more genres as well as translation pairs would be helpful in creating more robust MT systems. This will provide resources that are more suitable for drawing more reliable insights. Moreover, the challenges outlined in this paper can be subject of future research for development of better performing systems. Lastly, evaluation metrics can be explored in order to determine the most suitable ones for evaluation of Kurdish MT systems.

**ACKNOWLEDGMENTS**

We would like to thank the following individuals for providing us with resources or advice throughout the composition of the paper: Kourosh Abdi, Dr. Alan Dilani, Aso Mahmudi, Dr. Manijeh Mirmokri, Abdulkhaleq Yaqubi, Davoud Osmanzadeh, Shafiq Haji Khedr, Kosar Abdulla, Huma Hiwa, and Halo Fariq.